\documentclass[a4paper,10pt,twocolumn]{article}  

\usepackage{geometry}
\usepackage{authblk}
\usepackage[pdftex]{graphicx} 
\usepackage{titlesec}
\usepackage[labelfont=bf]{caption}
\usepackage{enumitem}
\usepackage{hyperref}
\usepackage{helvet}
\usepackage{etoolbox}
\usepackage{setspace}  
\usepackage{tabularx}
\usepackage{multirow}
\usepackage[export]{adjustbox}
\usepackage{microtype}
\usepackage{mathrsfs}
\usepackage{color}

\usepackage[skip=2pt]{caption}
\DeclareCaptionFormat{myformat}{\fontsize{9}{9}\selectfont#1#2#3}
\usepackage{etoolbox}
\patchcmd{\thebibliography}
{\settowidth}
{\setlength{\parsep}{0pt}\setlength{\itemsep}{0pt plus 0.1pt}\settowidth}
{}{}
  
\titleformat{\section}
{\normalfont\fontsize{10}{0}\bfseries}{\thesection}{0em}{}
\titlespacing\section{0pt}{12pt }{3pt }
\makeatletter
\patchcmd{\@maketitle}{  \@title}{\fontsize{14}{0}\selectfont\sffamily \bfseries \@title}{}{}
\makeatother

\title{Abdominal Aortic Aneurysm Segmentation \\with a Small Number of Training Subjects \vspace{-0.4cm}}

\author[1]{Jian-Qing Zheng}
\author[1]{Xiao-Yun Zhou}
\author[1]{Qing-Biao Li}
\author[2,3]{Celia Riga}
\author[1]{Guang-Zhong Yang \vspace{-0.3cm}}

\affil[1]{The Hamlyn Centre for Robotic Surgery, Imperial College London, UK}
\affil[2]{Academic Division of Surgery, Imperial College London, UK}
\affil[2]{Regional Vascular Unit, St Mary’s Hospital, London, UK}
\affil[ ]{ {j.zheng17@imperial.ac.uk} \vspace{-0.5cm}}

\begin{document}
	
	\date{}
	\maketitle
	
	\thispagestyle{empty}
	\pagestyle{empty}
	
		\begin{spacing}{0.6}
			\textit{\textbf{Abstract}} - Pre-operative Abdominal Aortic Aneurysm (AAA) 3D shape is critical for customized stent graft design in Fenestrated Endovascular Aortic Repair (FEVAR). Traditional segmentation approaches implement expert-designed feature extractors while recent deep neural networks extract features automatically with multiple non-linear modules. Usually, a large training dataset is essential for applying deep learning on AAA segmentation. In this paper, the AAA was segmented using U-net with a small number (two) of training subjects. Firstly, Computed Tomography Angiography (CTA) slices were augmented with gray value variation and translation to avoid the overfitting caused by the small number of training subjects. Then, U-net was trained to segment the AAA. Dice Similarity Coefficients (DSCs) over 0.8 were achieved on the testing subjects. The PLZ, DLZ and aortic branches are all reconstructed reasonably, which will facilitate stent graft customization and help shape instantiation for intra-operative surgery navigation in FEVAR.
	        \par
	        \textbf{\textit{Index Terms}} - AAA segmentation, U-net, data augmentation and FEVAR.
	
		\end{spacing}

	\section*{INTRODUCTION}
\begin{figure*}[ht]
		\centering
		\includegraphics[width=0.95\textwidth]{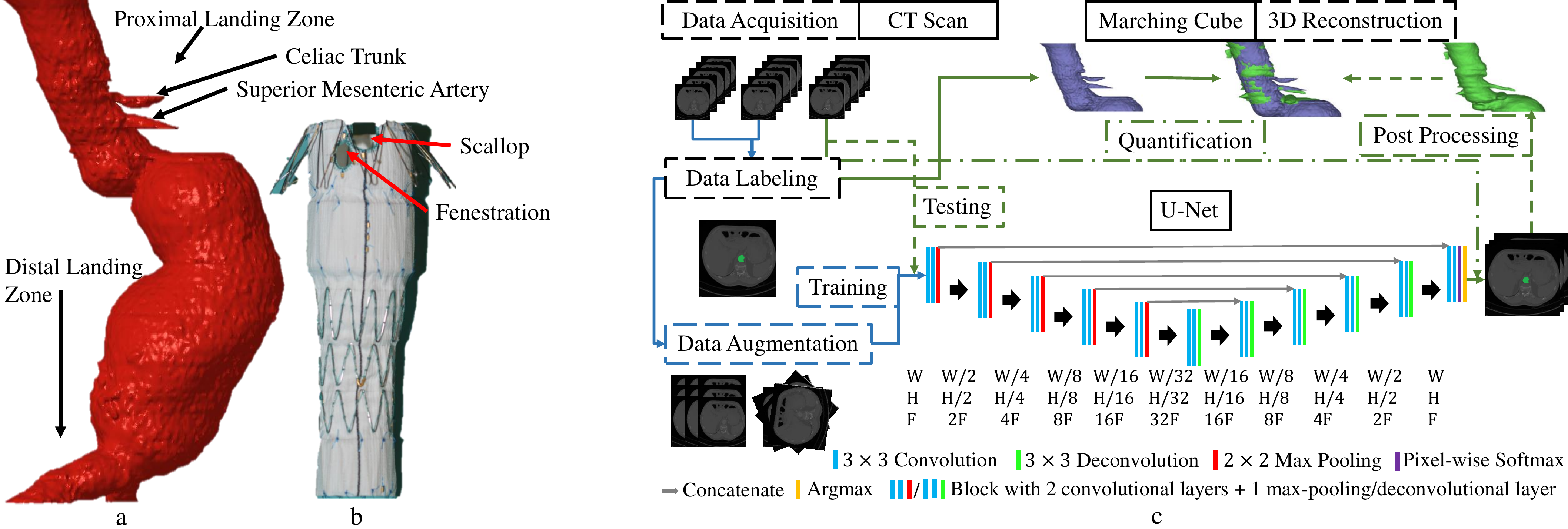}
		\vspace*{-0.4em}
		\caption{ \small (a) a 3D mesh model of one AAA, showing the PLZ, DLZ, etc.; (b) a main stent graft with customized fenestration and scallop; (c) a work flow chart of 3D AAA reconstruction, the 34-layer U-net with a convolution stride as 1, input width $\textrm{W}=512$, input height $\textrm{H}=512$, and initial feature number $\textrm{F}=64$.}
		\label{fig:combine}
		\vspace{-0.8em}
	\end{figure*}

	Abdominal Aortic Aneurysm (AAA) mostly occurs among old people and potential rupture can be deadly. Fenestrated Endovascular Aortic Repair (FEVAR) is a common procedure for treating AAA when it involves aortic branches, where multiple customized stent grafts including the main, branch and iliac stent grafts are usually cascaded to exclude the aneurysm. For stent graft customization, the diameters of the Proximal Landing Zone (PLZ) and the Distal Landing Zone (DLZ) determine the top diameter of the main stent graft and the bottom one of the iliac stent graft; the position and orientation of the aortic branch, including the celiac trunk and the superior mesenteric artery, determine those of the fenestration and scallop. A 3D AAA shape and its corresponding main stent graft are respectively shown in Fig.~\ref{fig:combine}(a) and Fig.~\ref{fig:combine}(b).
	\par
	AAA segmentation, indicating the shapes of PLZ, DLZ and aortic branch, is vital to stent graft customization. Traditionally, expert-designed feature extractors were implemented to segment the AAA while deep neural networks with multiple non-linear modules were recently used for automatic feature extraction. For example, a deep belief network was utilized to detect, segment, and classify the AAA \cite{hong2016automatic}, however, without quantified segmentation results provided. A deep convolutional neural network (DCNN)-based segmentation and quantification of abdominal aortic thrombus \cite{lopez2017dcnn} was implemented in totally 38 pre- and post-operative subjects to assess the EVAR treatments \cite{lopez2017dcnn}. A multi-stage network including a detection network (holistically-nested edge detection network) and a segmentation network (fully convolutional network) was proposed to segment the thrombi from 13 post-operative CTA volumes \cite{lopez2018fully}.

	\par
	Many training subjects were required in previous work. This paper caters for the overfitting caused by the small number of training subjects using data augmentation based on gray value variation and translation (G.\&T.) for training a one-stage network - U-Net \cite{ronneberger2015u}. The methodology workflow illustrated in Fig.~\ref{fig:combine}(c) includes data collection, data augmentation, network training, post-processing and 3D reconstruction. Dice Similarity Coefficients (DSCs) over 0.8 were achieved.
	
	\section*{MATERIALS AND METHODS}
	\par
	\textit{\textbf{Network Architecture and Training:}} U-net \cite{ronneberger2015u} with convolutional layers, max-pooling layers, and deconvolutional layers was adopted to train and segment the AAA slice-by-slice. Its detailed structure is shown in Fig.~\ref{fig:combine}(c). The input is a 2D CTA slice while the output is the segmentation result. Cross-entropy was computed as the loss function:
	\begin{center}
		\vspace*{-0.4em}
		${\mathcal{L}}_{CE}=-\sum _{i=1}^{W}\sum _{j=1}^{H}\sum _{k=1}^{N}\textit{\textbf{G}}_{(i,j,k))}\textrm{log}(\textit{\textbf{P}}_{(i,j,k)})~~(1)$
		\vspace*{-1.55em}
	\end{center}
	\par
	Here, $\textit{\textbf{G}}$ is the ground truth;  $\textit{\textbf{P}}$ is the predicted probability; W is the image width; H is the image height; $N=2$ is the class number.
	\par
	Stochastic Gradient Descent (SGD) and Adam were employed to minimize the loss function with the batch size as $1$ and no batch accumulation. For SGD, the initial learning rate was $0.1$ and multiplied by $0.1$ when the loss stops decreasing. A momentum of $0.9$ was applied. The initial learning rate of Adam was $0.001$. The loss converged after about $110k$ iterations.
	\par
	\textit{\textbf{Data Collection:}} Three pre-operative AAA CTA volumes were collected in a supine position from the celiac trunk to the iliac arteries from different patients. Two of them (Subject 2, Subject 3) scanned by Siemens Definition AS+ were contrast-enhanced while the other one (Subject 1) scanned by Philips ICT 256 was not. The pixel spacing was 0.645-0.977mm and the slice thickness was 0.7-1mm.  The AAA in the transverse sectional slices were manually segmented as the ground truth using \textit{Analyze} (AnalyzeDirect, Inc, Overland Park, KS, USA). All images were adjusted with a unified pixel spacing of $0.645\times0.645mm$ by sampling or bilinear interpolation.
	\par
	\textit{\textbf{Data Augmentation:}} A small number of training subjects results in overfitting due to the inter-subject variations in the CTA slice contrast (different contrast media used) and the AAA position. In this paper, the inter-subject variation of image contrast was compensated using a linear mapping transformation, and that of AAA position was solved by translating a $512\times512$ image window along the CTA slices with a stride of 64. We also applied rotation and mirroring (R.\&M.) data augmentation as a comparison, which was usually used with sufficient training subjects. The data were augmented 45-72 times from hundreds of original slices.
	\par
	\textit{\textbf{Post-processing and 3D Reconstruction:}} $\textit{MATLAB}^{\textregistered}$ functions "regionprops" and "bwareopen" were used to automatically extract the largest volume from the segmentation results. The 3D AAA shape was reconstructed using marching cube.
	\par
	\textit{\textbf{Validation:}} Cross-validation was adopted, in which respectively each subject was used for the testing while the other two for the training, except for Subject 1, as it was scanned without contrast media. DSC was computed for each slice to quantify the segmentation results, and \textit{Cloudcompare} was used to evaluate the 3D AAA reconstruction.
	\begin{table}[ht]
		\centering
		\caption{\small DSCs of AAA segmentation with cross-validation, different data augmentation, and U-Net with varying layers (Num.-Number).}
		\vspace*{-0.3em}
		\label{tab:result}
		\begin{tabular}{ccccc}
			\hline
			Row&Train & Augmen- & Layer & DSC \\
			&Subject & tation & Num.    & avg.$\pm$std\\\hline
			1&$1,2$   & R.\&M.    & 34 & $0.500\pm0.304$ \\ 
			2&$1,2$    & G.\&T.    & 28 & $0.757\pm0.151$ \\ 
			3&$1,2$    & G.\&T.    & 34 & $0.802\pm0.157$ \\ 
			4&$1,3$    & G.\&T.    & 34 & $0.824\pm0.131$ \\\hline
		\end{tabular}
		\vspace{-1.0em}
	\end{table}
	
	\begin{figure}[ht]
		\centering
		\includegraphics[width=0.48\textwidth]{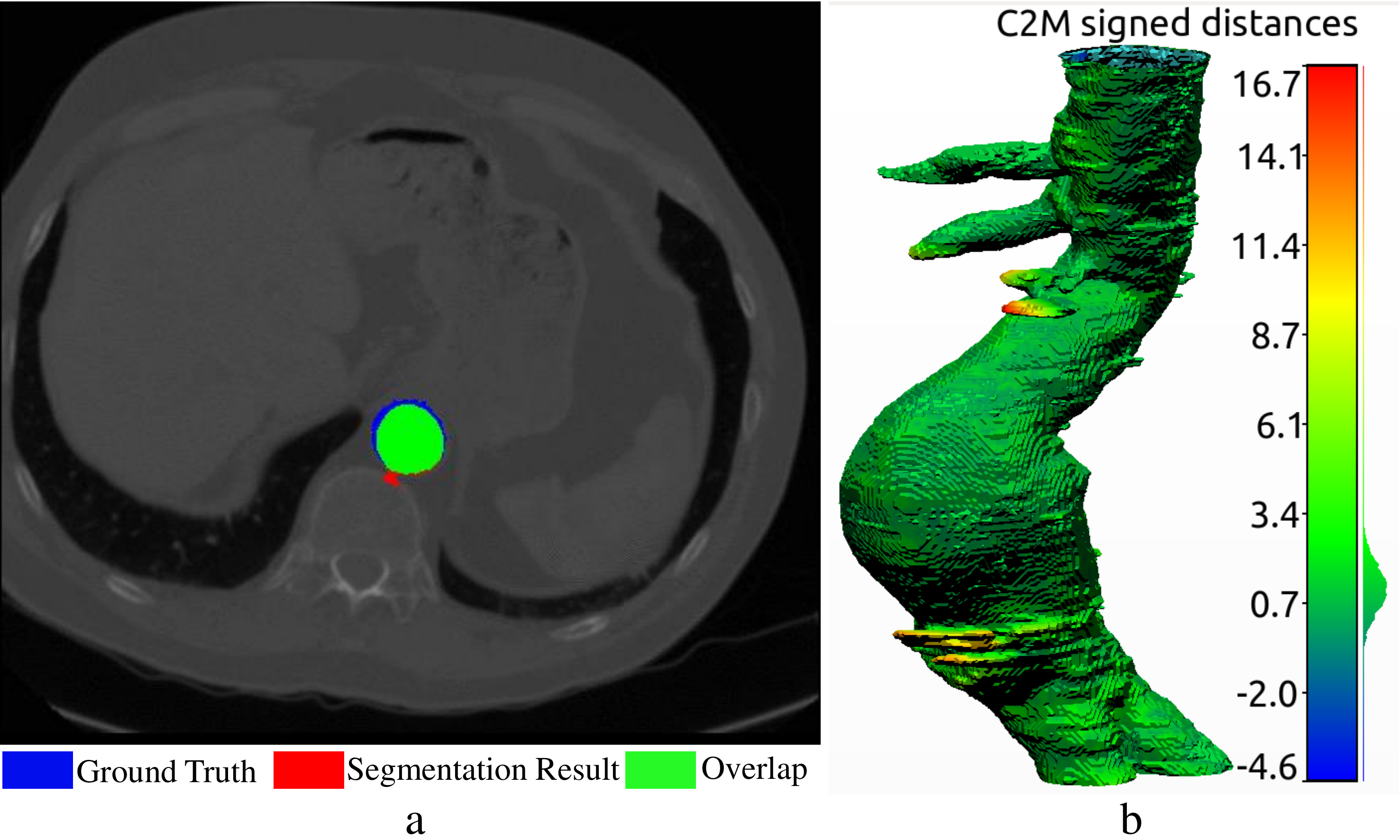}
		\vspace*{-1.6em}
		\caption{\small (a) one segmentation example; (b) a 3D outcome model, with the colours showing the point cloud-to-mesh distances (C2M, unit: pixel) between the model and the ground truth calculated using iterative closest point, and a histogram of C2M distance-distribution.}
		\label{fig:result}
		\vspace{-0.4em} 
	\end{figure}
	
	\section*{RESULTS}
	The U-Net was trained on different subjects, with varying layers and augmentation methods, the corresponding DSCs are shown in Tab.~\ref{tab:result}. Row 1 and Row 2 illustrate that the proposed data augmentation with G.\&T. outperformed the conventional R.\&M.; Row 2 and Row 3 show that a 34-layer U-Net achieved a higher DSC than a 28-layer U-Net; Row 3 and Row 4 verify the robustness of the proposed method to variable datasets. A segmentation example is demonstrated in Fig.~\ref{fig:result}(a), and a 3D reconstruction result is illustrated in Fig.~\ref{fig:result}(b), showing that the PLZ, DLZ and aortic branch are all reconstructed reasonably.
	
	\section*{DISCUSSION}
	In this paper, data augmentation G.\&T. was proposed to reduce the requirement of training subject number to two. We achieved one-stage AAA segmentation with U-Net from pre-operative CTA images, which facilitates stent graft customization and will help intra-operative FEVAR navigation in the future.

	\bibliographystyle{ieeetr}
	\small
	\bibliography{reference}

\end{document}